\documentclass[10pt,twocolumn,letterpaper]{article}

\usepackage{cvpr}              

\definecolor{cell}{RGB}{230, 237, 233}

\usepackage{graphicx}
\usepackage{booktabs}
\usepackage{adjustbox}
\usepackage{tabularx}
\usepackage{bbm}
\usepackage{wrapfig}
\usepackage{amssymb}
\usepackage{amsmath}
\usepackage{multirow}
\usepackage{array}

\definecolor{cvprblue}{rgb}{0.21,0.49,0.74}
\usepackage[pagebackref,breaklinks,colorlinks,allcolors=cvprblue]{hyperref}

\def\paperID{2943} 
\def\confName{CVPR}
\def\confYear{2025}

\title{IAAO: Interactive Affordance Learning for Articulated Objects in 3D Environments}

\author{Can Zhang \qquad Gim Hee Lee\\
Department of Computer Science, National University of Singapore\\
{\tt\small can.zhang@u.nus.edu \qquad gimhee.lee@nus.edu.sg}
}

\begin{document}
\maketitle
\begin{abstract}

This work presents IAAO, a novel framework that builds an explicit 3D model for intelligent agents to gain understanding of articulated objects in their environment through interaction. 
Unlike prior methods that rely on task-specific networks and assumptions about movable parts, our IAAO leverages large foundation models to estimate interactive affordances and part articulations in three stages. 
We first build hierarchical features and label fields for each object state using 3D Gaussian Splatting (3DGS) by distilling mask features and view-consistent labels from multi-view images. We then perform object- and part-level queries on the 3D Gaussian primitives to identify static and articulated elements, estimating global transformations and local articulation parameters along with affordances.
Finally, scenes from different states are merged and refined based on the estimated transformations, enabling robust affordance-based interaction and manipulation of objects. Experimental results demonstrate the effectiveness of our method. 
Our source code is available at: \url{https://lulusindazc.github.io/IAAOproject/}.

\end{abstract}    
\section{Introduction}
\label{sec:intro}

\begin{figure}[t]
\centering
\noindent\makebox[0.45\textwidth][c]{\includegraphics[scale=0.3]{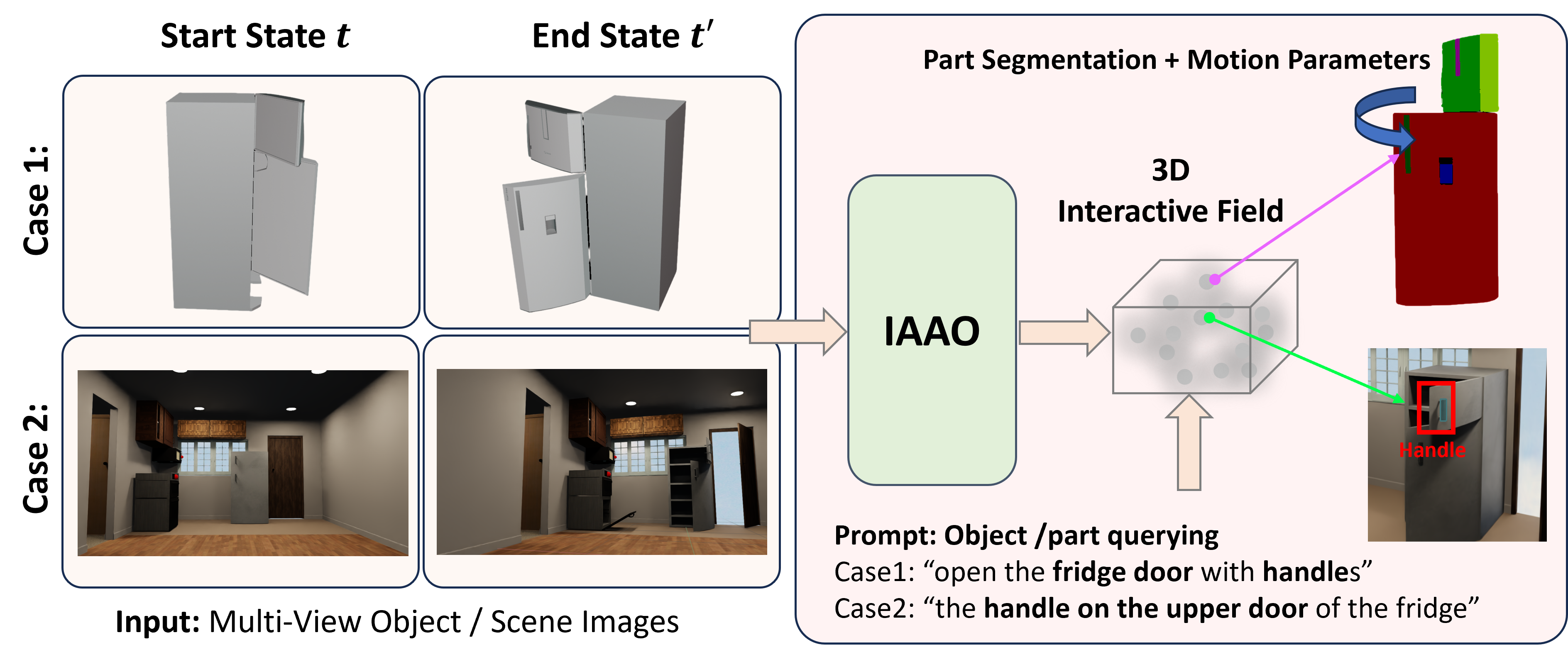}}
\caption{Our IAAO requires multi-view images of the object or indoor scene from two different joint states (Left). The output is a 3D interactive field which supports interactions with multiple movable parts for fine-grained segmentation (\eg Case 2: handles) and articulation reconstruction (\eg Case 1: two articulated doors). }
\label{fig:teaser}
\end{figure}

The real world contains various dynamic objects that continuously interact and move through time. Understanding and reconstructing the shapes, poses, and articulations of dynamic objects is crucial for many applications such as robotics, augmented reality and virtual reality (AR/VR), \etc While humans naturally develop an intuitive understanding of how objects move and interact, teaching this skill to an intelligent agent remains a complex challenge due to the diverse and intricate nature of articulated objects.
Furthermore, a key capability still missing in intelligent agents is effective interaction with their environment, particularly with functional elements like door handles and light switches. For intelligent agents to interact with functional objects, in addition to object detection, they must also understand the specific physical interactions each object affords—known as affordance~\cite{gibson2014ecological}. The small size of these elements further intensifies the challenge, making them difficult to detect. 
In this work, we aim to reconstruct the 3D complex scene and interact with both objects with static geometry and objects with articulations and fine-grained affordances as illustrated in Fig.~\ref{fig:teaser}.

Early approaches~\cite{wang2019shape2motion, li2020category} to understanding articulated objects rely on supervised learning with costly 3D annotations and category-specific models, which limit generalization to novel objects. Recent methods~\cite{jiang2022ditto,liu2023paris} such as Ditto~\cite{jiang2022ditto} and PARIS~\cite{liu2023paris} have advanced this field by creating digital twins of objects based on observations in two joint configurations. Since Ditto is trained on multi-view point clouds from specific categories, it struggles with objects that differ significantly from its training data. In contrast, PARIS~\cite{liu2023paris} uses multi-view posed images to directly optimize an implicit representation without pretraining, allowing for improved generalization. However, it remains sensitive to initialization and may experience stability issues. 
Further developments~\cite{weng2024neural,swaminathan2025leia} have expanded scalability to objects with multiple moving parts based on implicit neural radiance fields (NeRFs). 
Nonetheless, several challenges remain unexplored, including generalizing to complex scenes, flexibly applied for manipulation, and practical assumptions (\eg handling unknown camera poses, unaligned static object parts, and occlusion).


In 3D perception, instance segmentation provides essential object-level understanding for agents interacting within scenes. Recent studies~\cite{mo2019partnet, mao2022multiscan} explore part-object segmentation to decompose objects into finer components such as cabinet drawers. However, these methods are preliminary and struggle with affordance detection due to the lack of precise small-scale geometry in existing datasets. 
Open-vocabulary models~\cite{engelmannopennerf, kerr2023lerf, peng2023openscene, takmaz2024openmask3d} utilize foundation models such as CLIP~\cite{radford2021learning} to extract semantic features and consolidate them into a multi-view consistent 3D neural field representation. However, they rely on costly multi-scale queries to refine segmentation boundaries, and face challenges in high-dimensional feature encoding. 

To address the 
limitations mentioned above, we introduce IAAO, a framework designed for interaction of intelligent agents with objects, parts, and functional affordances, as well as for articulating movable objects. We explicitly leverage 3D semantic information to provide our model with scene understanding and interaction capabilities for articulated objects. Key insights of our approach include:~1) semantic scene reconstruction where SAM masks enable multi-view consistent segmentations from partial views, 2) hierarchical feature fields to efficiently compress dense features from multiple views into 3DGS for object- and part-level interaction in diverse scenes, 3) explicit 3D Gaussian representations for affordance localization via query similarity and motion recovery via 3D point-to-2D pixel correspondences, 4) global-local transformation initialization to enhance robustness in realistic scenarios with potential scene misalignment between object states, and 5) scene state fusion merging two different scene states to enhance object geometry completeness. 

Specifically, our IAAO enables interactive affordance detection and object articulation via the following steps: 1) \textbf{\textit{3D model construction}.} Using multi-view posed images of each state, we model the 3D scene as a set of explicit 3D Gaussian primitives with 3D Gaussian Splatting (3DGS)~\cite{kerbl20233d}. 
2) \textbf{\textit{Hierarchical feature field construction}.} We propose an efficient distillation scheme where shape-aware 2D features and dense semantic features are extracted from large foundation models, including CLIP~\cite{radford2021learning}, SAM~\cite{kirillov2023segany}, and DINOv2~\cite{oquab2023dinov2}. 2D features are efficiently distilled into 3D fields with a decoder. 3) \textbf{\textit{Semantic-guided mask association across states}.} We first cluster SAM-generated masks from all views within each state to obtain view-consistent mask labels. 
We then compute 3D proposal features by merging the features from the corresponding 2D masks. By comparing pairwise feature similarities, we establish 3D mask-level correspondences across states that result in consistent mask label sets for the entire scene in each state. 
4) \textbf{\textit{Affordance prediction}.} Given a task description, we encode it with the CLIP text encoder and compare the text embeddings with the feature fields to identify the relevant affordance. 5) \textbf{\textit{Motion recovery}.} To estimate transformations, we define a set of consistency and matching losses that recover both global scene-level and local part-level transformations that ensure accurate motion representation across different states.

We summarize our \textbf{main contributions} as follows:
\begin{itemize}
\item We introduce IAAO, an interactive affordance system utilizing 3D Gaussian fields embedded with hierarchical language-aligned semantics and class-agnostic masks. This system supports manipulation tasks guided by various prompts (point, mask, and language) at both object and part levels, regardless of object categories and shape.
\item We propose a method to reconstruct motion via global matching at the scene level and local matching for articulated objects, without relying on impractical assumptions about static object alignment or known camera poses.
\item Our IAAO achieves state-of-the-art performance across extensive experiments on multiple benchmarks, including synthetic, real-world, and indoor scene data. 
\item We show strong model generalization to complex indoor environments and previously unseen articulated objects, with no restrictions on the number of movable parts. 
\end{itemize}

\section{Related Work}
\label{sec:related_work}

\noindent \textbf{Neural Fields for 3D Scene Understanding.}
3D Scene understanding is primarily characterized by four classic representations, including volumetric fields (\cite{paris2006surface}), point clouds (\cite{schonberger2016structure}), 3D meshes (\cite{esteban2004silhouette}), and depth maps (\cite{goesele2006multi, strecha2006combined, yao2018mvsnet}). Unlike traditional representations, NeRF (Neural Radiance Fields)~\cite{mildenhall2021nerf} introduces a neural implicit field to capture the geometry and appearance of a scene. Through a Multi-Layer Perceptron (MLP) that takes 3D positions and 2D view directions as inputs, NeRF learns to implicitly represent the color and radiance of the scenes from a collection of posed images. However, a key drawback of NeRF is the slow training and rendering speed. Recently, 3D Gaussian Splatting (3DGS) \cite{kerbl20233d} has been introduced as an alternative to implicit radiance field representations. In contrast to NeRF, 3DGS explicitly represents radiance fields as a collection of oriented 3D Gaussians. Each Gaussian is defined by its spatial position, opacity, and a covariance matrix, which allows flexible optimization. With efficient differentiable rasterization, 3DGS achieves fast rendering with high-quality results. Various work leverage neural fields as a representation robotic manipulation~\cite{kerr2023lerf, wi2022virdo, zhou2023nerf, tang2023rgb, zhu2021rgb, wang2022clip}. Methods that leverage visual foundation models to construct neural feature fields~\cite{kerr2023lerf, qin2024langsplat, zhou2024feature} are most relevant to our work. With feature distillation via rendering, we reconstruct a hierarchical feature field for object localization and fine-grained affordance prediction.

\noindent \textbf{Visual Affordance.}
Affordance~\cite{gibson1977theory} refers to the potential interactions that an object or its parts facilitate for an agent. Affordance prediction involves deducing interaction opportunities from visual representations, \eg images~\cite{do2018affordancenet, fang2018demo2vec, luddecke2017learning} or 3D models~\cite{mo2022o2o, nagarajan2020learning, deng20213d}. This field has gained significant interest in robotics as a foundational element for tasks \eg grasping~\cite{mandikal2021learning, zeng2022robotic}, planning~\cite{xu2021deep} and exploration~\cite{nagarajan2020learning}. However, although these methods excel at predicting affordance regions, they lack detailed interaction information, which is essential for engaging with functional elements. Recent work~\cite{delitzas2024scenefun3d} addresses fine-grained functional elements using comprehensive natural language task descriptions for interaction. However, it depends on high-fidelity point clouds, restricting its practical applicability. 
We use high-resolution images with a Vision-Language Model (VLM) and 3D motion heuristics to predict affordances, eliminating the need for high-fidelity point clouds.

\noindent \textbf{Articulation Reasoning by Interaction.}
As embodied AI continues to evolve, the study of articulated objects has become a critical area of research. Recently, several synthetic~\cite{wang2019shape2motion, xiang2020sapien} and scanned data~\cite{jiang2022opd, liu2022akb, mao2022multiscan, qian2022understanding} datasets have been introduced for articulated objects. These datasets contain annotations for part segmentation and motion parameters that enable data-driven approaches in the prediction of motion parameters from 3D point clouds~\cite{wang2019shape2motion, yan2020rpm}. Recent works have been increasingly focused on real-world scenarios by detecting articulated parts and the motion parameters from images~\cite{zeng2021visual, jiang2022opd} and videos~\cite{qian2022understanding}. Interactive perception involves agents gaining new insights from their interactions with the environment, and has been widely used to study object articulations~\cite{nie2022structure, hsu2023ditto, jiang2022ditto, hausman2015active}. In this work, we focus on using language, and geometric and semantic scene understanding to achieve articulation reasoning.

\section{Our Method}
\label{sec:method}

\noindent \textbf{Objective.} 
Given a scene of articulated objects represented by a set of multi-view posed images $\mathcal{I}^t=\{I^t_i\}_{i=1}^{N^t}$ at two states $t$ and $t'$, the goal is to reconstruct the geometries and semantics of the objects in the scene. We focus on articulated objects which contain one or more movable parts with rotary or prismatic joints. Under the condition of unknown object catergory in the scene, we aim to recover the object part geometry, segmentation and joint articulations in the presence of occlusions, shape discrepancies and motion changes across different scene states. For static objects in the scene, the task degenerates to global transformation estimation between two input scene states for state matching.

\begin{figure*}[t]
\centering
\noindent\makebox[1\textwidth][c]{\includegraphics[scale=0.35]{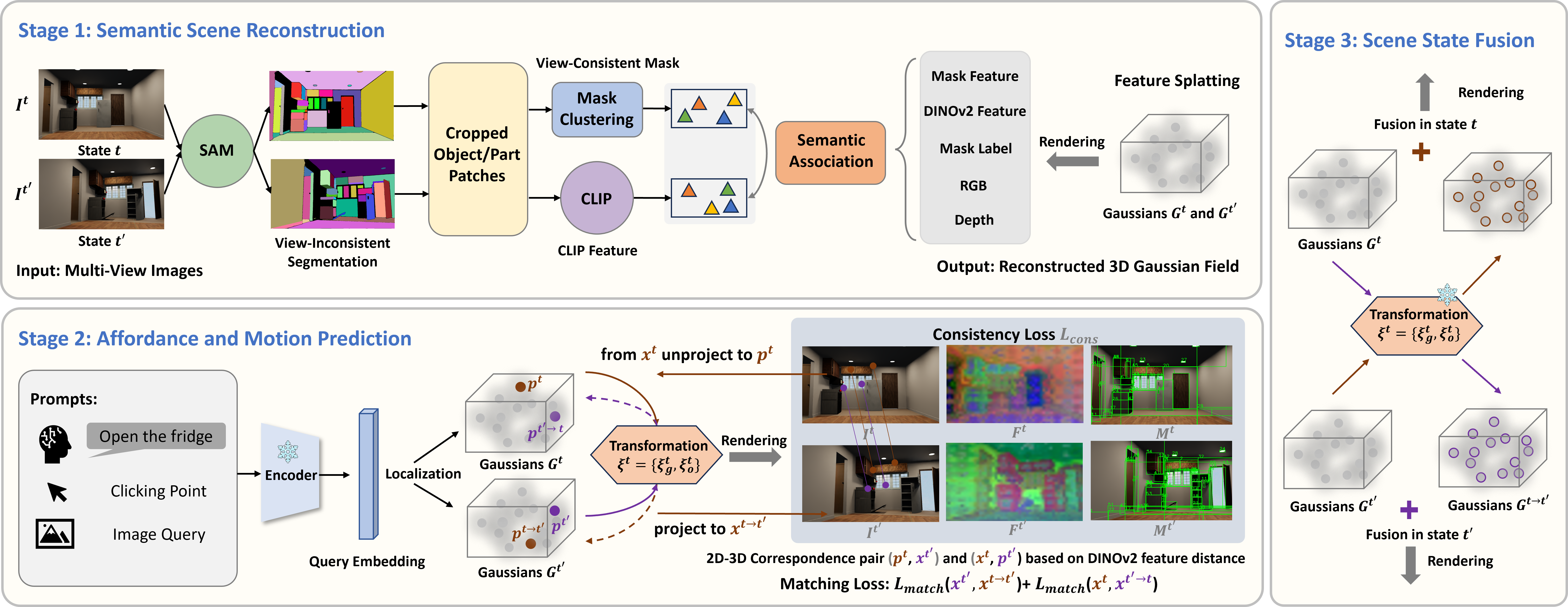}}
\caption{ \textbf{Our IAAO framework.} 1) Top: Constructing 3D Gaussian fields in each state. We optimize 3DGS fields with hierarchical mask features, DINOv2 features and 3D-consistent mask labels generated from multi-view images. We also incorporate geometry information from depth images into the 3D Gaussians.  
2) Bottom: Affordance and motion prediction. A query prompt is embedded using a pretrained encoder to localize relevant regions in the 3D Gaussians. For motion prediction, we optimize the transformation parameters by applying consistency and matching losses to 2D-3D correspondences between states.  
3) Right: Scene fusion. Using the estimated transformations, we merge reconstructed 3DGS models from both states, aligning static and articulated elements.}
\label{fig:framework}
\end{figure*}

\vspace{1mm}
\noindent \textbf{Overview.} Fig.~\ref{fig:framework} shows our IAAO framework consisting of: 1) A \textbf{\textit{semantic scene reconstruction}} stage (\cf Sec.~\ref{sec:semanticRecon}) where we first extract CLIP features for objects in individual scene state. These features are regularized by class-agnostic masks generated from 2D foundation models, \eg SAM~\cite{mobile_sam} at the instance and part level. We then cluster view-inconsistent masks from all views to obtain 3D consistent labels, and build a hierarchical feature field and label field for all masks in each scene state via 3D Gaussian Splatting (3DGS). 2) An \textbf{\textit{affordance and motion prediction}} stage (\cf Sec.~\ref{sec:affordMotionPred}) where we perform object- and part-level queries by directly operating on 3D primitives represented by 3D Gaussians. Static elements and articulated objects are identified by isolating the groups of 3D Gaussians corresponding to segmented objects and functional parts of the scene, \ie affordance prediction. For motion recovery, we estimate global transformation from static Gaussian primitives and local articulation parameters from the segmented movable parts of objects. 3) A \textbf{\textit{scene state fusion}} stage (\cf Sec.~\ref{sec:sceneFusion}) enabling affordance interaction and object manipulation by merging and refining the reconstructed scenes from two states according to the estimated transformations. 

\subsection{Semantic Scene Reconstruction}
\label{sec:semanticRecon}
Given a multi-view capture $\mathcal{I}^t$ of scene state $t$, we construct a 3D model using 3DGS initialized with sparse point cloud generated from Structure from Motion (SfM)~\cite{schonberger2016structure}. Each point serves as the center of a Gaussian primitive embedded with geometry and appearance parameters. These 3D Gaussian primitives $G^t={g^t_p}_{p=1}^{P^t}$ are then rendered into 2D views via differentiable rasterization for parameter optimization. To enrich the reconstructed model with semantic information, we augment each 3D Gaussian with semantic feature embeddings derived from large foundation models.

\vspace{1mm}
\noindent \textbf{View-Consistent Mask Clustering.}
For each input image $I^t_i \in \mathcal{I}^t$, we first derive the class-agnostic masks $M^t_i=\{m_{i,n}^t\}_{n=1}^{n^t_i}$  with an off-the-shelf mask predictor, where $n^t_i$ represents the number of masks in view $i$. For $M^t_i$, we investigated the use of SAM~\cite{mobile_sam} to generate both instance- and part-level masks for arbitrary objects. Unfortunately, mask generation models such as SAM face several challenges: a) class-agnostic masks do not have complete 3D information, b) inconsistent segmentation across images due to variations in viewpoint and appearance, and c) lack of one-to-one correspondence among masks due to over-segmentation for individual object. We aim to generate 3D-consistent labels for all masks in $M^t$ by defining a mask label mapping function $\phi(M^t,G^t) \mapsto O^t$. This function maps a set of masks \( M^t_o \) belonging to the same 3D object to a consistent $o^t\in O^t$ within the 3D Gaussian primitives $G^t$.

To associate class-agnostic masks across all input views at scene state \( t \), we construct a mask graph to fuse the 2D masks $M^t$ of the entire scene into cohesive 3D instances $O^t$. In the initial graph \( \mathcal{G}^t_0 = (\mathcal{V}^t_0, \mathcal{E}^t_0) \), each node in \( \mathcal{V}^t_0 \) represents a mask detected from the views, and each edge in \( \mathcal{E}^t_0 \) represents potential matches between pairs of object parts. Following the mask clustering workflow from~\cite{yan2024maskclustering}, we then cluster the mask nodes and update the edges to facilitate mask association across different views. For each cluster, we combine the corresponding partial sets of Gaussians from individual masks to form a complete 3D instance and establish correspondences between 2D masks and 3D nodes. This results in the final graph \( \mathcal{G}^t = (\mathcal{V}^t, \mathcal{E}^t) \), where each node in $\mathcal{V}^t $ represents a 3D instance proposal in $O^t$ in scene state \( t \). We maintain a list of associated 2D masks \( M^t_o \) for each $o^t $, discarding severely occluded masks. This 2D-3D relationship enables consistent segmentation labels across all views. However, the mask label mapping \( \phi(M^t, G^t) \) may be affected by noise due to varying part segmentation hierarchies across views. To mitigate segmentation ambiguities, we use the labeled masks \( M^t_o \) in the final graph to supervise a label field for object part $o^t$ in the given scene state. To this end, we use a cross-entropy classification loss \( \mathcal{L}_{\text{label}} \) for enhanced consistency.

\noindent \textbf{Per-State Hierarchical Feature Field.}
We generate a set of feature embeddings from the given image \( I^t_i \) using pretrained foundation models. Specifically, we utilize MaskCLIP~\cite{dong2023maskclip} to produce mask-level CLIP features at both instance and part levels, and DINOv2~\cite{oquab2023dinov2, darcet2023vitneedreg} to extract DINO embeddings. To enable object- and part-level interaction, we construct a hierarchical feature field by embedding dense features from multi-view images into the 3D Gaussian primitives. Using semantic hierarchy from SAM, we enhance our 3D semantic field accuracy while making the querying process more efficient. Directly embedding high-dimensional 2D features would result in excessive computational and memory costs. Therefore, we build a low-dimensional latent feature field and introduce a decoder to project the rendered features \( \hat{F}^t \) back to the 2D feature space $F^{t}$ during the differentiable rasterization optimization process as: 
\begin{equation}
    \mathcal{L}_{feat}= \parallel \mathcal{D}(\hat{F}^t(I^t_i)) - F^{t}(I^t_i) \parallel_2^2.
    \label{eq:render_feat}
\end{equation}
The decoder \( \mathcal{D} \) is implemented as a small MLP with three output branches: two for instance- and part-level CLIP features, and one for DINO features.

\noindent \textbf{Semantic Association between Scene States.}
Upon obtaining the mask label mapping functions $O^t$ and $O^{t'}$ for scene states $t$ and $t'$, we establish a global 3D mask association which links the same semantic regions across both states into a unified 3D mask using their semantic features. For each 3D proposal in $O^t$, we select the top-$k$ representative masks from $M^t_o$ and aggregate their rendered features to create the corresponding semantic primitives $F^t_o$. We then build an affinity matrix capturing the cosine similarity between all 3D proposal pairs $(O^t, O^{t'})$ using their primitives $(F^t_O,F^{t'}_O)$ from two scene states. For each query proposal $o^t \in O^t$, we define its visual match as the target proposal in $O^{t'}$ with the highest similarity.

\subsection{Affordance and Motion Prediction}
\label{sec:affordMotionPred}
Unlike NeRF-based methods~\cite{kerr2023lerf, engelmannopennerf, liu2023weakly} which require resource-intensive rendering to derive language-embeded features and geometry from implicit MLPs, our IAAO directly uses explicit Gaussian primitives for efficient object part localization and manipulation. Given the reconstructed 3D scene $G^t=\{g^t_p\}_{p=1}^{P^t}$ at scene state $t$, where each Gaussian point $g^t_p$ is embedded with geometric information (coordinates of points $p^t$) and additional features (color and semantic embeddings), our goal is to enable scene-aware affordance prediction and object articulation estimation.

\noindent \textbf{Functional Affordance Prediction.}
Functional affordances are defined as parts of the scene that facilitate interactions for agents to perform specific tasks. We utilize the reconstructed hierarchical feature and label fields as the 3D representations to predict the masks $\{m_a\}_{a=1}^{K_a}$ and affordance labels $\{l_a\}_{a=1}^{K_a}$ in the scene, where $K_a$ is the total number of affordance categories. We first build associations between affordance labels and our reconstructed mask label field, and then infer the point-level affordance of a functional element in the scene. Additionally, for task-specific reasoning, we encode the task description using the CLIP text encoder. To identify the relevant mask for a given description, we calculate the similarity between the embeddings in the feature field and the query embeddings. A mask is retrieved if its similarity score with the description text exceeds a predefined threshold.

\noindent \textbf{Global and Local Motion Recovery.}
Given the reconstructed scenes with different object articulations at two states, our objective is to estimate scene-aware transformation and 3D motion primitives for articulated parts. 
Our approach segments the reconstructed Gaussians $G^t$ and $G^{t'}$ into static scene elements and articulated objects via \textit{\textbf{object-part querying}} for global and per-part local transformation estimation, respectively. 
1) \textbf{\textit{Object-level querying}} in complex scenes: In an indoor scene with a complex background and multiple objects, we start with object-level querying. Positive language queries are used to identify target objects for manipulation, while optional negative queries exclude irrelevant items. 
This selection process identifies 3D Gaussians with object-level CLIP features that align more closely with the positive queries than the negative ones, following standard CLIP feature-text comparison practices~\cite{wang2022clip}. 
2) \textbf{\textit{Part-level querying}} for targeted objects: For individual objects in the scene, we apply part-level querying to focus on 3D Gaussians within the object. This allows for fine-grained transformation estimation specific to each articulated part. In addition to language, other user input prompts such as masks and points can also be used. Nonetheless, we do not further elaborate on these prompts as they can be easily obtained by a user clicking on the images.

We define the \textit{\textbf{global scene-aware transformation}} between two different scene states as $\xi_g^t=(s_g^t, R_g^t, T_g^t)$, where $s_g^t \in \mathbbm{R}$ represents the scale factor, $R_g^t \in \operatorname{SO}(3)$ is the rotation matrix and $T_g^t \in \mathbbm{R}^{3}$ is the translation vector. To initialize the transformation, we estimate an initial alignment using 3D Gaussians derived from static elements in the scene, excluding articulated objects. 
From the Gaussian points in both scene states, we select reliable points with opacity values above a threshold (empirically set at 0.7) to serve as input point sets for the coarse registration. Using the GeoTransformer~\cite{qin2022geometric}, we calculate the scene alignment transformation by matching coarse superpoints and refining with dense point correspondences. 
Since these point clouds are based on reconstructed 3D Gaussians, the resulting transformation is coarse, providing an initial approximation for $\xi_g^t$ that can be further refined.

After achieving global alignment between two scene states, we define the \textit{\textbf{local transformation}} for each articulated part as $\xi_o^t=(R_o^t, T_o^t)$, representing the movement of part $o$ from state $t$ to state $t'$. All points within a part are assumed to follow the same motion parameters. The reverse mapping, which transforms part $o$ from state $t'$ back to state $t$, is given by the inverse transformation, $\xi_o^{t'}=(R_o^t, T_o^t)^{-1}$. To facilitate this process, we construct a 3D Gaussian correspondence field, allowing each Gaussian primitive $g^t_p(o)$ at state $t$ to be repositioned to its corresponding location $p^{t'}$ at state $t'$ based on the motion parameters of part $o^t$. Formally: 
\begin{equation}
\centering
\begin{split}
    & p^{t\rightarrow t'}=
    \left\{
    \begin{array}{ll}
    f(p^t, \xi_g^t) = R_g^t p^t + T_g^t, 
    & \mbox{if \textit{p} in static scene},  \\
    f(f(p^t, \xi_g^t), \xi_o^t),  & \mbox{otherwise.}
    \end{array}
    \right. \\
\end{split}
\label{eq: pseudo-label}
\end{equation}
The same rule applies to the backward transformation function $f^{-1}$ for obtaining $p^{t'\rightarrow t}$ (\ie mapping from $t'$ back to $t$) with the transformations $\xi_g^{t'}$ and $\xi_o^{t'}$.

Establishing accurate dense correspondences between point clouds in two states is challenging due to the sparsity and noise in 3D points derived from the 3DGS model. To optimize the correspondence field, we instead match 3D Gaussian primitives to 2D pixels across the two states, guided by part geometry and segmentation information. 
Specifically, for a Gaussian primitive $g^t_p(o)$ of part $o$ at state $t$, the corresponding 2D pixels at state $t'$ can be treated as observations from novel views. 
To enhance consistency, we incorporate mask-level features alongside the rendered image consistency loss $\mathcal{L}_{rgb}$. 
For a 3D proposal $o^t$ in graph $\mathcal{G}^t$, we identify the part mask list $\mathcal{M}^{t'}_o$ associated with the corresponding proposal $o^{t'}$ at state $t'$. Following the splatting process, we render the feature field of $o^t$ to training view $n$ in target scene state $t'$ and decode the mask-level features by $\mathcal{D}(\hat{F}^{t}_o (I^{t'}_n))$. We formulate the mask feature loss for part $o$ transformed from state $t$ to target state $t'$ as:
\begin{equation}
\begin{split}
    & \mathcal{L}_{mask}(o^{t\rightarrow t'}) = \\
    & \Sigma_{n}^{N_o^{t'}} \frac{M^{t'}_o(I^{t'}_n)}{\mid M^{t'}_o(I^{t'}_n)\mid} \cdot \parallel \mathcal{D}(\hat{F}^{t}_o (I^{t'}_n))- F^{t'}_{o}(I^{t'}_n)\parallel_2^2,
\end{split}
\end{equation}
where $M^{t'}_o(I^{t'}_n)$ and $F^{t'}_{o}(I^{t'}_n)$ represent the corresponding part mask and 2D mask-level features of part $o$ in view $I^{t'}_n$ at the target state $t'$. $N_o^t$ denotes the number of part masks for proposal $o^t$ across all training views.

Since the mask-level loss provides only coarse guidance for transformation, we further propose to find dense correspondences between the 3D Gaussian primitives and 2D pixels by comparing their DINO features. 
We begin by computing the feature similarity matrix \( \alpha_{p\rightarrow o}(I^{t'}_n) \) between each pixel in the part mask $M^{t'}_o(I^{t'}_n)$ in view $I^{t'}_n$ and the sampled Gaussian point \( g^t_p\) from 3D proposal $o^t$. We then normalize the similarity matrix \( \alpha_{p\rightarrow o}(I^{t'}_n) \) using a softmax over the entire mask to obtain the weight \( \beta_{p\rightarrow o}(I^{t'}_n) \). Finally, we identify the corresponding 2D pixel \( s_{p \rightarrow o}^{t'}(I^{t'}_n) \) for the sampled 3D point $ p^t $ using a weighted sum. 
See the supplementary material for details on the computation.
For each point-pixel pair \( (p^t, s_{p \rightarrow o}^{t'}(I^{t'}_n)) \), we filter out pairs that collide by checking whether \( s_{p \rightarrow o}^{t'}(I^{t'}_n) \) falls within the same part mask \( o^{t'} \). We then define the matching loss as follows:
\begin{equation}
     \mathcal{L}_{match}= 
     \parallel \pi^{t'}_n(p^{t\rightarrow t'}) - s_{p\rightarrow n}^{t'}(I^{t'}_n) \parallel_2^2,
\end{equation}
where $\pi^{t'}_n$ represents the projection of 3D point $p^{t\rightarrow t'}$ to view $I^{t'}_n$ in scene state $t'$.

\noindent \textbf{Optimization.}
For the first stage, each scene is reconstructed independently at two different states. We optionally use depth map for further geometry regularization. 
In the second stage, we reconstruct motion of global scene and local articulated objects. The total loss is defined as: 
\begin{equation}
\mathcal{L} = \lambda_{cons} (\mathcal{L}_{rgb}+\mathcal{L}_{mask}+\mathcal{L}_{label}) + \lambda_{match} \mathcal{L}_{match}.
\end{equation}
$\lambda_{cons}$ and $\lambda_{match}$ denote the balancing weights of consistency loss and matching loss.

\subsection{Scene State Fusion}
\label{sec:sceneFusion} 
After training, we merge the 3DGS models from the two states using the estimated transformations to fill in the occluded regions in each state. Details of our fusion and filtering strategy are provided in the supplementary material. The combined 3D Gaussians facilitate object manipulation by addressing potential artifacts that may appear in occluded areas of the reconstructed scene. Optional fine-tuning can be applied to either the edited regions or the entire scene for further refinement.

\section{Experiment}
\label{sec:experiment}


\subsection{Experimental Setup}
\noindent \textbf{Datasets.}
1) \textit{\textbf{PARIS two-part object dataset.}} This dataset contains two-part articulated objects, featuring 10 synthetic objects from {PartNet-Mobility~\cite{xiang2020sapien} and 2 real-world objects captured using MultiScan~\cite{mao2022multiscan}. Each object comprises a movable part and a static part, observed in two distinct joint states. The dataset includes RGB images, object masks from 100 random viewpoints, and depth images for both synthetic and real-world objects. 
2) \textit{\textbf{Synthetic multi-part object dataset.}} This dataset includes two synthetic scenes containing multi-part objects from PartNet-Mobility, with each object featuring one static part and multiple movable parts. Each object is observed in two articulation states, and the dataset provides RGB, depth, and mask data from 100 random viewpoints. 
3) \textit{\textbf{Indoor scene OmniSim dataset.}} 
OmniSim is generated using the OmniGibson~\cite{li2024behavior} simulator with various indoor scene models. By adjusting the rotation of articulated object joints, we generate some scenes for evaluation, each containing RGBD images, interactive object masks and object state metrics at each state.

\vspace{1mm}
\noindent \textbf{Metrics.}
To evaluate \textit{\textbf{articulation models}}, we use: 1) \textit{Axis Ang Err} $(^\circ)$ means angular error of the predicted joint axis for both revolute and prismatic joints. 2) \textit{Axis Pos Err} (0.1 m): The shortest distance between the predicted and true joint axes for revolute joints.
For \textit{\textbf{cross-state part motion}}, we use \textit{Part Motion Err} (° or m). It evaluates joint state precision by calculating the geodesic error for predicted rotations in revolute joints or the Euclidean error for translations in prismatic joints, as specified in PARIS~\cite{liu2023paris}.
For \textit{\textbf{object and part mesh reconstruction}}, we use Chamfer-L1 distance (CD), reporting CD-w for the full surface, CD-s for static parts, and CD-m for movable parts. 

\noindent \textbf{Baselines.}
\textit{\textbf{Ditto}}~\cite{jiang2022ditto} and \textit{\textbf{PARIS}}~\cite{liu2023paris} are models for reconstructing part-level shape and motion in two-part articulated objects, using multi-view data. Ditto assumes one moving part and is pre-trained on specific categories, while PARIS uses a NeRF-based representation to handle unknown objects. \textit{\textbf{PARIS*}} is an enhanced version with depth supervision, and \textit{\textbf{PARIS-m*}} extends PARIS to multi-part objects. \textit{\textbf{CSG-reg}} uses TSDF fusion and Constructive Solid Geometry for part segmentation, with registration for alignment. \textit{\textbf{3Dseg-reg}} follows a similar process but employs PA-Conv~\cite{xu2021paconv} for part segmentation, reporting results only for trained categories due to limited generalization.
\textit{\textbf{DigitalTwinArt}}~\cite{weng2024neural} divides reconstruction into two stages: 
1) It reconstructs object-level shape independently of articulation; 2) It recovers the articulation model by identifying part segmentation and motions through state correspondences.

\begin{table*}[th]
    \centering
    \begin{adjustbox}{width=1\textwidth}
    \begin{tabular}{llcccccccccccccc}
        \hline
        \multicolumn{2}{l}{\multirow{2}{*}{}} & \multicolumn{11}{c}{\textbf{Simulation} } &\multicolumn{3}{c}{\textbf{Real} } \\
        & & \textbf{FoldChair} & \textbf{Fridge} & \textbf{Laptop} & \textbf{Oven} & \textbf{Scissor} & \textbf{Stapler} & \textbf{USB} & \textbf{Washer} & \textbf{Blade} & \textbf{Storage} & \textbf{All} & \textbf{Fridge} & \textbf{Storage} & \textbf{All} \\ \hline
        \multirow{6}{*}{\textbf{Axis Ang $\downarrow$}} & Ditto~\cite{jiang2022ditto} & 89.35 & 89.78 & 3.12 & 0.96 & 4.50 & 89.86 & 89.77 & 89.51 & 79.54* & 6.32 & 54.22 & \textbf{1.71} & \textbf{5.88} & \textbf{3.80} \\ 
        & PARIS~\cite{liu2023paris} & 8.08  & 9.15  & \textbf{0.02} & \textbf{0.04} &3.82  & 39.73  & 0.13 & 25.36  & 15.38 & \textbf{0.03} & 10.17 & 1.64 & 43.13  & 22.39  \\ 
        & PARIS*~\cite{liu2023paris} & 15.79  & 2.93  & 0.03  & 7.43 & 16.62  & 8.17  & 0.71 & 18.40 & 41.28 & \textbf{0.03} &11.14 & 1.90 &30.10&16.00 \\ 
        & CSG-reg & 0.10  & 0.27 & 0.47  & 0.35 & 0.28  & 0.30  &  11.78  &71.93& 7.64  & 2.82  & 9.60 & 8.92 & 69.71 & 39.31  \\ 
        & 3Dseg-reg &- & -  & 2.34 & - & -  & - & -  &- & 9.40 & - &-&-&-&-  \\ 
        & DigitalTwinArt~\cite{weng2024neural}& 0.03 & 0.07 & 0.06  & 0.22  & 0.11  & 0.06 &0.11 & 0.43  & 0.27  & 0.06 &0.14 & 2.10  & 18.11 & 10.11 \\  
         & \cellcolor{cell}IAAO (Ours) & \cellcolor{cell}\textbf{0.02} & \cellcolor{cell}\textbf{0.04} & \cellcolor{cell}0.05 & 
         \cellcolor{cell}0.17 &  \cellcolor{cell}\textbf{0.08} &   \cellcolor{cell}\textbf{0.05} &  \cellcolor{cell}\textbf{0.09} & \cellcolor{cell}\textbf{0.36} &  \cellcolor{cell}\textbf{0.19} &  \cellcolor{cell}0.06 &  \cellcolor{cell}\textbf{0.11} &
         \cellcolor{cell}1.9  & 
         \cellcolor{cell}15.82 & 
         \cellcolor{cell}8.86 \\ \midrule
         \multirow{5}{*}{\textbf{Axis Pos $\downarrow$}} 
        & Ditto~\cite{jiang2022ditto} & 3.77 & 1.02* & 0.01 & 0.13 & 5.70 & 0.20 & 5.41 & 0.66 & - & - & 2.11 & 1.84 & -& 0.92\\ 
        & PARIS~\cite{liu2023paris} & 0.45& 0.38& 0.00& 0.00 & 2.10 & 2.27& 2.36 & 1.50 & - & - & 1.13& 0.34 & -& 0.17 \\ 
        & PARIS*~\cite{liu2023paris}  & 0.25 & 1.13 & 0.00 & 0.05 & 1.59 & 4.67 & 3.35 & 3.28 & -& -  & 1.79& 0.50 & - & 0.25 \\ 
        & CSG-reg & 0.02 & 0.00& 0.20 & 0.18 & 0.01 & 0.02& 0.01 & 2.13 & -& -  &0.32 &1.46& -  & 0.73\\ 
        & 3Dseg-reg & - & - & 0.10&- & - & - & - & - & - & - & - & - & - & - \\ 
        &  DigitalTwinArt~\cite{weng2024neural}& \textbf{0.01} & 0.01 & 0.00 & 0.01 & 0.02 & 0.01 & 0.00 & 0.01 & - & - &\textbf{0.01}&  0.57& -  & 0.29\\ 
         & \cellcolor{cell} IAAO (Ours)& \cellcolor{cell} \textbf{0.01} & \cellcolor{cell} \textbf{0.00} & \cellcolor{cell}\textbf{0.00} &  \cellcolor{cell}\textbf{0.00} &  \cellcolor{cell}\textbf{0.01} & \cellcolor{cell}\textbf{0.01} & \cellcolor{cell}\textbf{0.01} &  \cellcolor{cell}\textbf{0.00} & \cellcolor{cell}-  & \cellcolor{cell}-  & \cellcolor{cell}\textbf{0.01}  & \cellcolor{cell}\textbf{0.32} & \cellcolor{cell}- & \cellcolor{cell}\textbf{0.16}\\ \midrule
        \multirow{5}{*}{\textbf{Part Motion $\downarrow$}} 
        & Ditto~\cite{jiang2022ditto} & 99.36 & $\times$ & 5.18 & 2.09 & 19.28 & 56.61 & 80.60 & 55.72 & $\times$ &0.09 &39.87 & 8.43 & 0.38 &4.41 \\ 
        & PARIS~\cite{liu2023paris} & 131.66& 24.58& \textbf{0.03} & 
        \textbf{0.03} & 
        120.70& 110.80 & 64.85 & 60.35 & 0.34 & 0.30&51.36 &2.16&0.56& 1.36\\ 
        & PARIS*~\cite{liu2023paris} & 127.34& 45.26& \textbf{0.03} & 9.13 & 68.36 & 107.76 & 96.93 & 49.77 & 0.36 &0.30& 50.52 &1.58&0.57& 1.07 \\ 
        & CSG-reg & \textbf{0.13} & 0.29& 0.35 & 0.58 & 0.20& 0.44 &10.48&158.99& 0.05 &0.04& 17.16& 14.82 & 0.64 &7.73 \\ 
        & 3Dseg-reg & - & - & 1.61  & - & - & - & - & - & 0.15 &- & - & - &-&-\\ 
        & DigitalTwinArt~\cite{weng2024neural} & 0.16 & 0.09& 0.08 & 0.11 & 0.15& \textbf{0.05} & 0.11& 0.25 & \textbf{0.00} & \textbf{0.00} &0.10& 1.86 & 0.20 & 1.03 \\  
        & \cellcolor{cell}IAAO (Ours)& 
        \cellcolor{cell}\textbf{0.13} & 
        \cellcolor{cell}\textbf{0.07} &  
        \cellcolor{cell}\textbf{0.03} & 
        \cellcolor{cell}0.04  &  
        \cellcolor{cell}\textbf{0.11} &  
        \cellcolor{cell}\textbf{0.05} & 
        \cellcolor{cell}\textbf{0.09} & 
        \cellcolor{cell}\textbf{0.21} & 
        \cellcolor{cell}\textbf{0.00} &  
        \cellcolor{cell}\textbf{0.00} & \cellcolor{cell}\textbf{0.07} &
        \cellcolor{cell}\textbf{1.46} &
        \cellcolor{cell}\textbf{0.15} &
        \cellcolor{cell}\textbf{0.81}  \\ \midrule
        \multirow{6}{*}{\textbf{CD-s $\downarrow$}} & Ditto~\cite{hsu2023ditto} & 33.79 & 3.05 & 0.25 & 2.52 & 39.07 & 41.64 & 2.64 & 10.32 & 46.90 & 9.18 & 18.94 & 47.01 & 16.09 & 31.55 \\
        &PARIS~\cite{liu2023paris} & 9.16 & 3.65 & 1.94 & 1.88 & 12.95 & 1.88 & 2.69 & 25.39 & 1.19 & 12.76 & 7.18 & 42.57 & 54.54 & 48.56 \\
        &PARIS*~\cite{liu2023paris} & 10.20 & 8.82 & 15.58 & 1.95 & 3.18 & 2.48 & 2.48 & 12.19 & 1.40 & 8.67 & 6.46 & 11.64 & 20.25 & 15.94 \\
        &CSG-reg & 1.69 & 1.45 & 3.93 & 3.26 & 3.26 & 2.22 & 1.95 & 4.53 & 0.59 & 7.06 & 2.70 & 6.33 & 12.55 & 9.44 \\
        &3Dseg-reg & - &-& 0.76 & - & - & - & - & - & 66.31 & - & - & - & - & -  \\
        &DigitalTwinArt~\cite{weng2024neural} & 0.18 & 0.60 & 4.66 & 0.40 & 2.65 & 2.19 & 4.80 & 4.69 & 0.55 & 4.69 & 2.10 & 2.53 & 10.86 & 6.69 \\
        & \cellcolor{cell}IAAO (Ours)  & 
        \cellcolor{cell}\textbf{0.16}  & 
        \cellcolor{cell}\textbf{0.51}  &  
        \cellcolor{cell}\textbf{0.17}  & 
        \cellcolor{cell}\textbf{0.36}  & 
        \cellcolor{cell}\textbf{1.98}  & 
        \cellcolor{cell}\textbf{1.86}  & 
        \cellcolor{cell}\textbf{4.43}  & 
        \cellcolor{cell}\textbf{3.70}  & 
        \cellcolor{cell}\textbf{0.51} &  
        \cellcolor{cell}\textbf{4.27} & 
        \cellcolor{cell}\textbf{1.80} &
        \cellcolor{cell}\textbf{2.47} &
        \cellcolor{cell}\textbf{9.83} &
        \cellcolor{cell}\textbf{6.15} \\ \midrule
        \multirow{6}{*}{\textbf{CD-m $\downarrow$}} &Ditto~\cite{hsu2023ditto} & 141.11 & 0.99 & 0.19 & 0.94 & 20.68 & 31.21 & 15.88 & 12.89 & 195.93 & 2.20 & 42.20 & 50.60 & 20.35 & 35.48 \\
        &PARIS~\cite{liu2023paris} & 8.99 & 7.76 & 0.21 & 28.70 & 46.64 & 19.27 & 5.32 & 178.43 & 25.21 & 76.69 & 39.72 & 45.66 & 864.82 & 455.24 \\
        &PARIS*~\cite{liu2023paris} & 17.97 & 7.23 & 0.15 & 6.54 & 16.65 & 30.46 & 10.17 & 265.27 & 117.99 & 52.34 &52.48&77.85 & 474.57 & 276.21 \\
        &CSG-reg & 1.91 & 21.71 & 0.42 & 256.99 & 1.95 & 6.36 & 29.78 & 436.42 & 26.62 & 1.39 &78.36 & 442.17 & 521.49 & 481.83  \\
        &3Dseg-reg & - &-& 1.01 & - & - & - & - & - & 6.23 & - & - & - & - & - \\
        &DigitalTwinArt~\cite{weng2024neural} & 0.15 & 0.27 & 0.16 & 0.47 & 0.41 & 2.27 & 1.34 & 0.36 & 1.50 & 0.37 & 0.73 & 1.14 & 26.46 & 13.80 \\
        & \cellcolor{cell}IAAO (Ours)& 
        \cellcolor{cell}\textbf{0.13}  & 
        \cellcolor{cell}\textbf{0.18}  &  
        \cellcolor{cell}\textbf{0.15}  & 
        \cellcolor{cell}\textbf{0.41}  & 
        \cellcolor{cell}\textbf{0.40}  & 
        \cellcolor{cell}\textbf{2.09}  &
        \cellcolor{cell}\textbf{1.18}  & 
        \cellcolor{cell}\textbf{0.33}  & 
        \cellcolor{cell}\textbf{1.42}  & 
        \cellcolor{cell}\textbf{0.25}  & 
        \cellcolor{cell}\textbf{0.65}  & 
        \cellcolor{cell}\textbf{1.02}  & 
        \cellcolor{cell}\textbf{21.43} & 
        \cellcolor{cell}\textbf{11.23} \\ \midrule
        \multirow{6}{*}{CD-w $\downarrow$} &Ditto~\cite{hsu2023ditto} &  6.80 & 2.16 & 0.31 & 2.51 & 1.70 & 2.38 & 2.09 & 7.29 & 42.04 & 3.91 & 7.12 & 6.50 & 14.08 & 10.29 \\
        &PARIS~\cite{liu2023paris} & 1.80 & 2.92 & 0.30 & 11.73 & 10.49 & 3.58 & 2.00 & 24.38 & 0.60 & 8.57 & 6.64 & 22.98 & 63.35 & 43.16 \\
        &PARIS*~\cite{liu2023paris} & 4.37 & 5.53 & 0.26 & 3.18 & 3.90 & 5.27 & 1.78 & 10.11 & 0.58 & 7.80 & 4.28 & 8.99 & 32.10 & 20.55 \\ 
        &CSG-reg & 0.48 & 0.98 & 0.40 & 3.00 & 1.70 & 1.99 & 1.20 & 4.48 & 0.56 & 4.00 &1.88& 5.71 & 14.29 & 10.00 \\
        &3Dseg-reg &  - &-& 0.81 & - & - & - & - & - & 0.78 & - & - & - & - & - \\
        &DigitalTwinArt~\cite{weng2024neural} & 0.27 & 0.70 & 0.35 & 4.18 & 0.43 & 2.19 & 1.18 & 4.74 & 0.36 & 3.99 & 1.84 & 2.19 & 9.33 & 5.76 \\
        & \cellcolor{cell}IAAO (Ours) &
        \cellcolor{cell}\textbf{0.22} & 
        \cellcolor{cell}\textbf{0.61} &  
        \cellcolor{cell}\textbf{0.27} & 
        \cellcolor{cell}\textbf{3.49} & 
        \cellcolor{cell}\textbf{0.34} &  
        \cellcolor{cell}\textbf{1.96} & 
        \cellcolor{cell}\textbf{1.06} & 
        \cellcolor{cell}\textbf{4.41} &  
        \cellcolor{cell}\textbf{0.34} &  
        \cellcolor{cell}\textbf{3.94} & 
        \cellcolor{cell}\textbf{1.66} & \cellcolor{cell}\textbf{2.02} & 
        \cellcolor{cell}\textbf{9.18} & 
        \cellcolor{cell}\textbf{5.60}  \\
        \bottomrule
    \end{tabular}
    \end{adjustbox} 
    \caption{Results on PARIS dataset including both synthetic and real data. `x': failure case, and `*': joint axis or position. '--': no result is available. Best result in \textbf{bold}.}
\label{tab:quantitative_paris}
\end{table*}


\begin{table*}[t!]
    \centering
    \begin{adjustbox}{width=1\textwidth}
    \begin{tabular}{clccccccccccc}
        \toprule
        & & \textbf{Axis Ang 0$\downarrow$}  & \textbf{Axis Ang 1 $\downarrow$} & \textbf{Axis Pos 0 $\downarrow$} & \textbf{Axis Pos 1 $\downarrow$} & \textbf{Part Motion 0 $\downarrow$} & \textbf{Part Motion 1 $\downarrow$} & \textbf{CD-s $\downarrow$} & \textbf{CD-m 0 $\downarrow$} & \textbf{CD-m 1 $\downarrow$} & \textbf{CD-w $\downarrow$} \\ \midrule
        \multirow{3}{*}{\textbf{Fridge-m}} & PARIS*-m~\cite{liu2023paris}& 34.52 & 15.91 & 3.60 & 1.63 & 86.21 & 105.86 & 8.52 & 526.19 & 160.86 & 15.00 \\ 
        & DigitalTwinArt~\cite{weng2024neural}  & \textbf{0.16} & 0.10 & 0.01 & 0.00 & 0.11 & 0.13 & 0.61 & \textbf{0.40} & 0.52 & \textbf{0.89} \\ 
         & \cellcolor{cell}IAAO (Ours) & 
         \cellcolor{cell}\textbf{0.16} & \cellcolor{cell}\textbf{0.08} &  \cellcolor{cell}\textbf{0.01} & \cellcolor{cell}\textbf{0.00} & \cellcolor{cell}\textbf{0.09} & 
         \cellcolor{cell}\textbf{0.09} & 
         \cellcolor{cell}\textbf{0.54} & 
         \cellcolor{cell}0.44 & 
         \cellcolor{cell}\textbf{0.47} &  
         \cellcolor{cell}0.96 \\ 
        \midrule
        \multirow{3}{*}{\textbf{Storage-m}} & PARIS*-m~\cite{liu2023paris} & 43.26 & 26.18 & 10.42 & - & 79.84 & 0.64 & 8.56 & 128.62 & 266.71 & 8.66 \\ 
        & DigitalTwinArt~\cite{weng2024neural}  & 0.21 & \textbf{0.88} & 0.05 & - & 0.13 & \textbf{0.00} & 0.85 & \textbf{0.21} & 3.46 & 0.99 \\ 
         & \cellcolor{cell}IAAO (Ours)  & 
         \cellcolor{cell}\textbf{0.18}  &  
         \cellcolor{cell}0.91 & 
         \cellcolor{cell}\textbf{0.04}  & 
         \cellcolor{cell} -  & 
         \cellcolor{cell}\textbf{0.10}  & 
         \cellcolor{cell}\textbf{0.00}  &  
         \cellcolor{cell}\textbf{0.78}  & 
         \cellcolor{cell}0.26  &  
         \cellcolor{cell}\textbf{3.35} &  
         \cellcolor{cell}\textbf{0.97} \\ 
        \bottomrule
    \end{tabular}
    \end{adjustbox} 
    \caption{ Results on multi-part object dataset, averaged over 10 trials with different random seeds. Joint 1 of “Storage-m” is solely prismatic with no Axis Pos. '--': no result is available. Best results in \textbf{bold}.} 
\label{tab:quantitative_multi}
\end{table*}


\noindent \textbf{Evaluation Setup.} 
There are some key differences on evaluation setup between our model and previous models as we firstly incorporate the semantics into the explicit neural field supporting robust interactions even in complex environment. 
Unlike prior methods which assume the number of parts is known, our model can detect any articulated objects if enough information for reconstructing object shape is given from multi-view images. 
To identify corresponding parts for evaluation, prior work need to iterate through all possible pairs of predicted and ground-truth parts, selecting the match with the smallest total Chamfer distance. In our model, we identify the corresponding parts based on the constructed semantic feature and mask label field. We extract the mesh based on the 3D point clouds derived from 3DGS model via SuGaR~\cite{guedon2024sugar} for evaluation. 
Following~\cite{liu2023paris, weng2024neural}, we transform our extracted parts
with predicted motions to start state $t$ for evaluation.



\begin{figure*}[t]
\centering
\noindent\makebox[1\textwidth][c]{\includegraphics[scale=0.55]{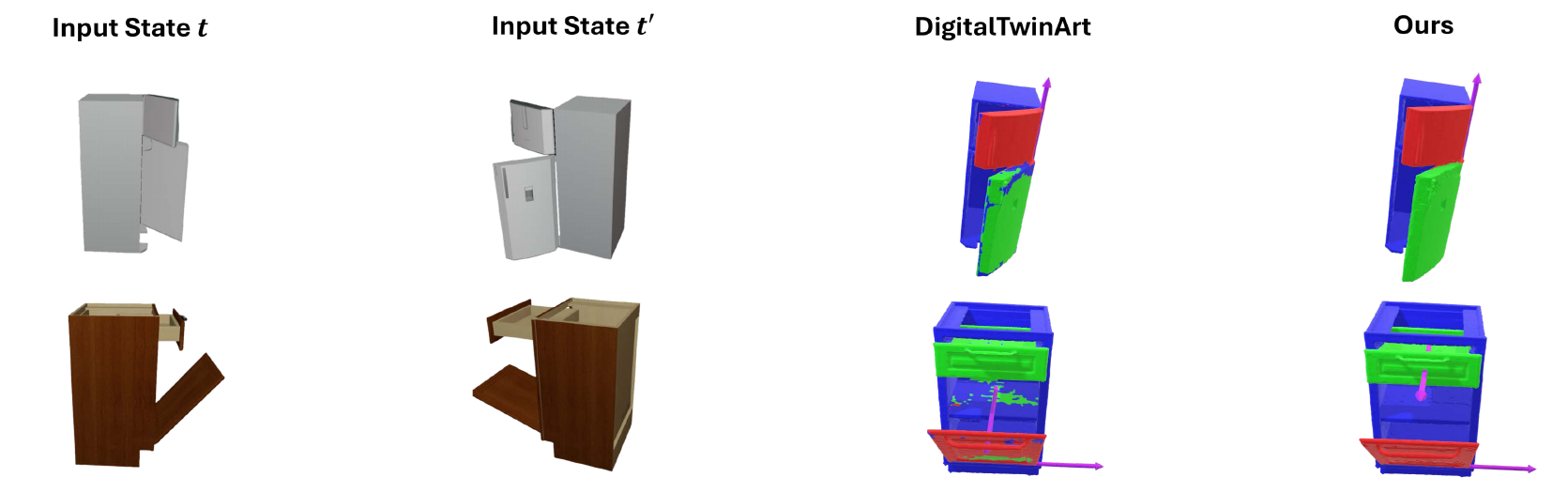}}
\vspace{-7mm}
\caption{Qualitative analysis of shape reconstruction, part segmentation, and joint prediction results on multi-part object dataset.}
\label{fig:qualitative_multi_part}
\end{figure*}

\begin{figure}[t!]
\centering
\noindent\makebox[0.5\textwidth][c]{\includegraphics[scale=0.3]{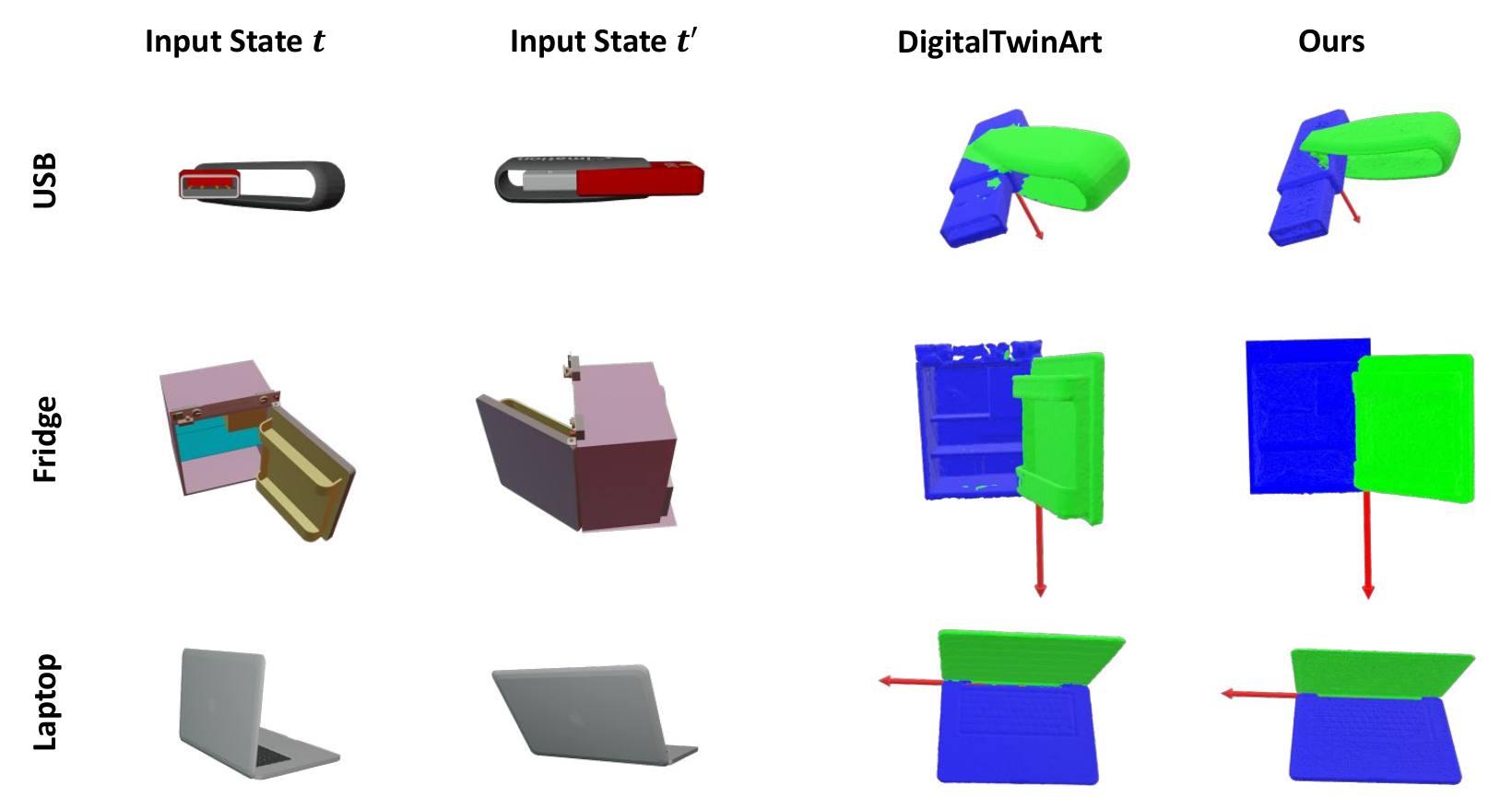}}
\vspace{-6mm}
\caption{Qualitative results of shape reconstruction, part segmentation, and joint prediction on PARIS.} 
\vspace{-3mm}
\label{fig:qualitative_paris}
\end{figure}

\begin{figure}[t] 
\centering
\noindent\makebox[0.5\textwidth][c]{\includegraphics[scale=0.3]{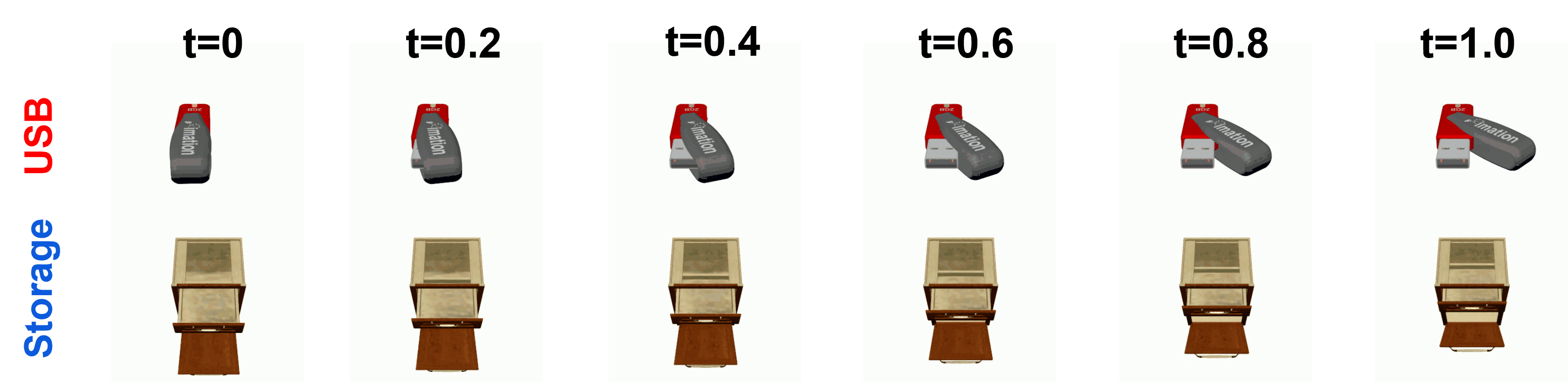}}
\vspace{-6mm}
\caption{Motion snapshots on \textcolor{red}{PARIS} \& \textcolor{blue}{multi-part object}.}
\label{fig:obj-level}
\vspace{-3mm}
\end{figure}

\begin{figure}[t]
\centering
\noindent\makebox[0.2\textwidth][c]{\includegraphics[scale=0.28]{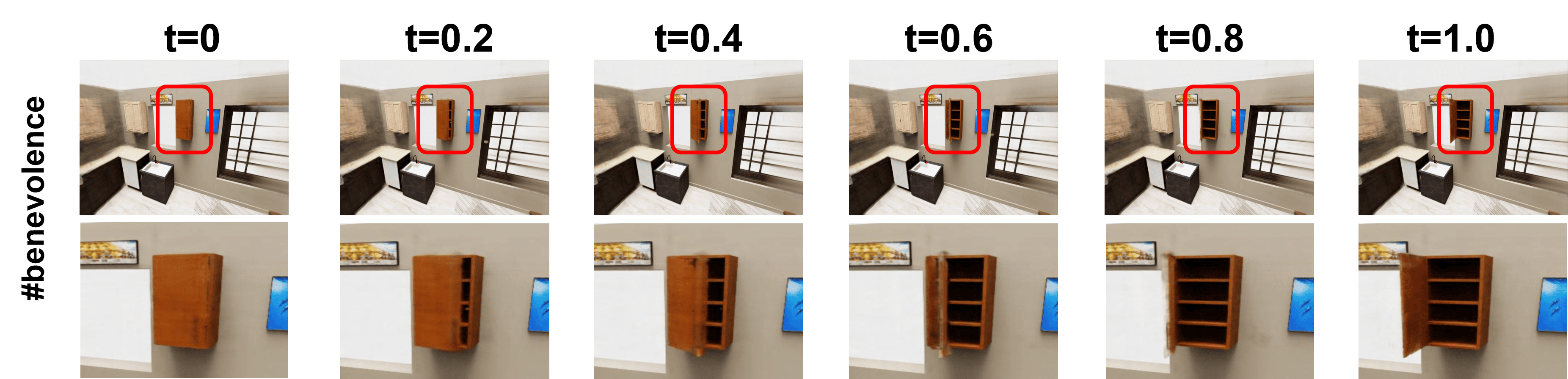}}
\vspace{-2mm}
\caption{Motion snapshots on scene-level OmniSim dataset.}
\label{fig:scene-level}
\vspace{-3mm}
\end{figure}

\subsection{Quantitative Results} 
Tab.~\ref{tab:quantitative_paris} shows quantitative results from the PARIS Two-Part Object Dataset, covering synthetic and real instances. 
Ditto performs well on seen categories but struggles with unseen ones. PARIS occasionally fails in shape and articulation reconstruction, while depth supervision in PARIS* improves shape accuracy for complex and real-world objects but increases articulation errors due to optimization challenges. CSG-reg and 3Dseg-reg handle simple objects well but struggle with complex ones, as segmentation errors affect articulation estimation.

Tab.~\ref{tab:quantitative_multi} presents the results for multi-part objects. 
DigitalTwinArt~\cite{weng2024neural} outperforms PARIS* in shape and articulation reconstruction by incorporating additional supervision on 3D geometries. Our IAAO achieves competitive results by leveraging geometric and semantic information from VLMs, enhancing object part modeling. 


\subsection{Qualitative Results} 

Fig.~\ref{fig:qualitative_multi_part} and Fig.~\ref{fig:qualitative_paris} show qualitative results on shape reconstruction, part segmentation, and joint prediction for the multi-part object and PARIS two-part datasets.
Our IAAO achieves more complete segmentation results while also enabling interaction with affordance and segmentation features. 
Fig.~\ref{fig:obj-level} and~\ref{fig:scene-level} show snapshots of the generated motions on several examples at object and scene levels in a simulator. It shows IAAO can produce smooth interpolations of articulated objects with good part geometry modeling. 




\subsection{Ablation Study}
We assess the impact of our design choices specifically on multi-part objects, which present greater challenges. We first perform experiments to test out the impact of 2D-3D macthing loss (`w/o matching'). As indicated in Tab.~\ref{tab:ablation}, the matching loss plays a crucial role in enhancing articulation reconstruction. We also study constraints from mask features and labels, which act as a coarse correspondence mechanism at the superpoint (mask) level. These constraints also enforce consistency within the feature fields of each mask, leading to improved part segmentation and more accurate reconstruction of articulated components.

\begin{table}[t]
\centering
\begin{adjustbox}{width=0.47\textwidth}
\begin{tabular}{lccccc}
\hline
\textbf{Method} & \textbf{Axis Ang $\downarrow$} & \textbf{Axis Pos $\downarrow$} & \textbf{Part Motion $\downarrow$} & \textbf{CD-s $\downarrow$} & \textbf{CD-m $\downarrow$} \\
\hline
Proposed       & 0.28 & 0.01 & 0.11 & 0.80 & 0.35 \\
w/o matching   & 4.59 & 1.38 & 8.57 & 1.21 & 123.71 \\
w/o mask     & 14.82 & 0.01 & 0.28 & 2.59 & 129.14 \\
w/o rgb        & 0.56 & 0.01 & 0.17 & 0.81 & 0.42 \\
w/o label        & 0.54 & 0.01 & 0.15 & 8.56 & 169.73 \\
\hline
\end{tabular}
\end{adjustbox} 
\caption{Ablation study on multi-part object dataset.} 
\vspace{-3mm}
\label{tab:ablation}
\end{table}
\section{Conclusion}
\label{sec:conclusion}
We present IAAO, a framework that facilitates interactive affordance detection and object articulation from two scene sequences. Each scene sequence captures movable parts of objects in different joint states. Our IAAO can handle both simple scenes with single objects and complex indoor scenes without limitations on the number of objects and movable parts. The entire scene is reconstructed using 3D Gaussians, embedded with robust zero-shot generalization capabilities from 2D foundation models. Affordance prediction is achieved through object and part queries on the 3D Gaussian primitives. 
Motion estimation is achieved by establishing 2D-3D correspondences within each object to track transformations. Static elements serve for global scene alignment while movable elements are for local transformation. Extensive experiments demonstrate that our approach outperforms existing methods.

\paragraph{Acknowledgement.}
This research / project is supported by the National Research Foundation (NRF) Singapore, under its NRF-Investigatorship Programme (Award ID. NRF-NRFI09-0008).

{
    \small
    \bibliographystyle{cvpr}
    \bibliography{cvpr}
}

\clearpage
\setcounter{page}{1}
\maketitlesupplementary


\section{Appendix}
\subsection{3D-2D Correspondence Matching}
We start with computing the feature similarity matrix \( \alpha_{ip} \) between each pixel in part mask $o$ and the sampled Gaussian point $p^t$. 
We then normalize similarity matrix \( \alpha_{ip} \) using a softmax across the entire mask to obtain the weight \( \beta_{ip} \). Finally, we identify the 2D point \( s_{p \rightarrow o}^{t'}(I^{t'}_n) \) corresponding to the 3D point using a weighted sum. 
The computation steps are as follows:
\begin{enumerate}
    \item Compute the feature distance:
    \[
    \alpha_{ip} = \| F^{t'}_{o}(I^{t'}_n) [u_i] - F^t_{3D,o}(p) \|_2,
    \]
    between the \( i \)-th pixel \( u_i \) of \( I_n^{t'} \) and sampled Gaussian point \( g^t_p (o) \).
    \item Normalize \( \alpha_{ip} \) using a softmax across the entire image to obtain the weight: 
    \[
    \beta_{ip} = \frac{\exp(-s \alpha_{ip})}{\sum_{i=1}^{\mid M^{t'}_o(n)\mid} \exp(-s \alpha_{ip})}.
    \]
    \item Identify the 2D point:
    \[
    s_{p \rightarrow o}^{t'}(I^{t'}_n) = \sum_{i=1}^{\mid M^{t'}_o(n)\mid} \beta_{ip} u_i,
    \]
    corresponding to the 3D point using a weighted sum.
\end{enumerate}
\( F^{t'}_{o}(I^{t'}_n)\) represents the DINOv2 features extracted from \(  I_n^{t'} \), and \( s \) is a hyperparameter that adjusts the smoothness of the heatmap \( \beta_{ij} \). 


\subsection{Scene State Fusion}
After obtaining the transformation $\xi^t=(s^t, R^t, T^t)$ (or its inverse function $\xi^{t'}=(s^{t'}, R^{t'}, T^{t'})$) for each part in scene state $t$, the next step is merging the two Gaussian Splatting (GS) models in the two states. We adopt the Gaussian splatting fusion and filtering strategy from~\cite{chang2024gaussreg}. To transform the Gaussians from the coordinate system of $G^{t'}=\{g^{t'}_p\}_{p=1}^{P^{t'}}$ to $G^t=\{g^t_p\}_{p=1}^{P^t}$, the position of each 3D Gaussian $g^{t'}_p$ is transformed as follows:
\[
(x^{t' \to t}_p, y^{t' \to t}_p, z^{t' \to t}_p)^\top = s^{t'} R^{t'} (x^{t'}_p, y^{t'}_p, z^{t'}_p)^\top + T^{t'}.
\]

The opacity remains unchanged during this transformation, i.e., $\alpha^{t' \to t}_p = \alpha^{t'}_p$. The rotation matrix $R^{t' \to t}_p \in \mathbb{R}^{3 \times 3}$ and scale $S^{t' \to t}_p \in \mathbb{R}^3$ are computed as:

\[
R^{t' \to t}_p = R^{t'} R^{t'}_p, \quad S^{t' \to t}_p = s^{t'} S^{t'}_p.
\]

Spherical harmonics (SH) coefficients undergo a linear transformation based on their rotation, which can be handled independently for each order. 
To this end, for any $i$-th order of SH coefficients, the following steps are performed: 

\begin{enumerate}
    \item Choose $2i+1$ unit vectors $u_0, \dots, u_{2i+1}$ and compute their corresponding SH coefficients as $Q = (\operatorname{SH}(u_0), \dots, \operatorname{SH}(u_{2i+1}))$.
    \item Apply the transformation $\xi^{t'}=(s^{t'}, R^{t'}, T^{t'})$ to the vectors $u_0, \dots, u_{2i+1}$ to obtain transformed vectors $\hat{u}_0, \dots, \hat{u}_{2i+1}$.
    \item Compute the transformation matrix for SH coefficients as:
    \[
    (\operatorname{SH}(\hat{u}_0), \dots, \operatorname{SH}(\hat{u}_{2i+1})) Q^{-1}.
    \]
\end{enumerate}


Finally, the 3D Gaussians in $G^t=\{g^t_p\}_{p=1}^{P^t}$ closer to the center of scene $t$ are merged with those in $G^{t'}=\{g^{t'}_p\}_{p=1}^{P^{t'}}$ near the center of scene $t'$, producing $G^{t+t'}$.

\subsection{More Visualization}

Fig.~\ref{fig:qualitative_paris2} and Fig.~\ref{fig:qualitative_paris3}
show the qualitative results on the ``PARIS Two-Part Object Dataset".
Column 4 shows the results of our IAAO and Column 3 shows the comparison with DigitalTwinArt. The input states at $t$ and $t'$ are shown in Columns 1 and 2. The part segments are shown in different colors and the red arrows indicate the joint prediction. Rotation is around the arrow and translation is along the arrow. Generally, as compared with DigitalTwinArt on both datasets, we can see from all figures that our IAAO produces better shape reconstruction with clearer details, more precise part segmentation with lesser erroneous labels, and more accurate joint predictions with correct motion directions. 
Fig.~\ref{fig:qualitative_paris_partnet}, \ref{fig:qualitative_omnisim1} and~\ref{fig:qualitative_omnisim2} show snapshots of the generated motions on several examples at object and scene levels in a simulator. It shows IAAO can produce smooth interpolations of articulated objects with good part geometry modeling. 
Fig~\ref{fig:qualitative_omni} illustrates the qualitative results on a sample scene from the Indoor Scene OmniSim dataset. Unlike existing baselines, which lack semantic meaning in their reconstructed neural fields, our model demonstrates the ability to perform object-level and fine-grained part localization based on prompts within complex indoor environments.



\begin{figure*}[th]
\centering
\noindent\makebox[1\textwidth][c]{\includegraphics[scale=0.55]{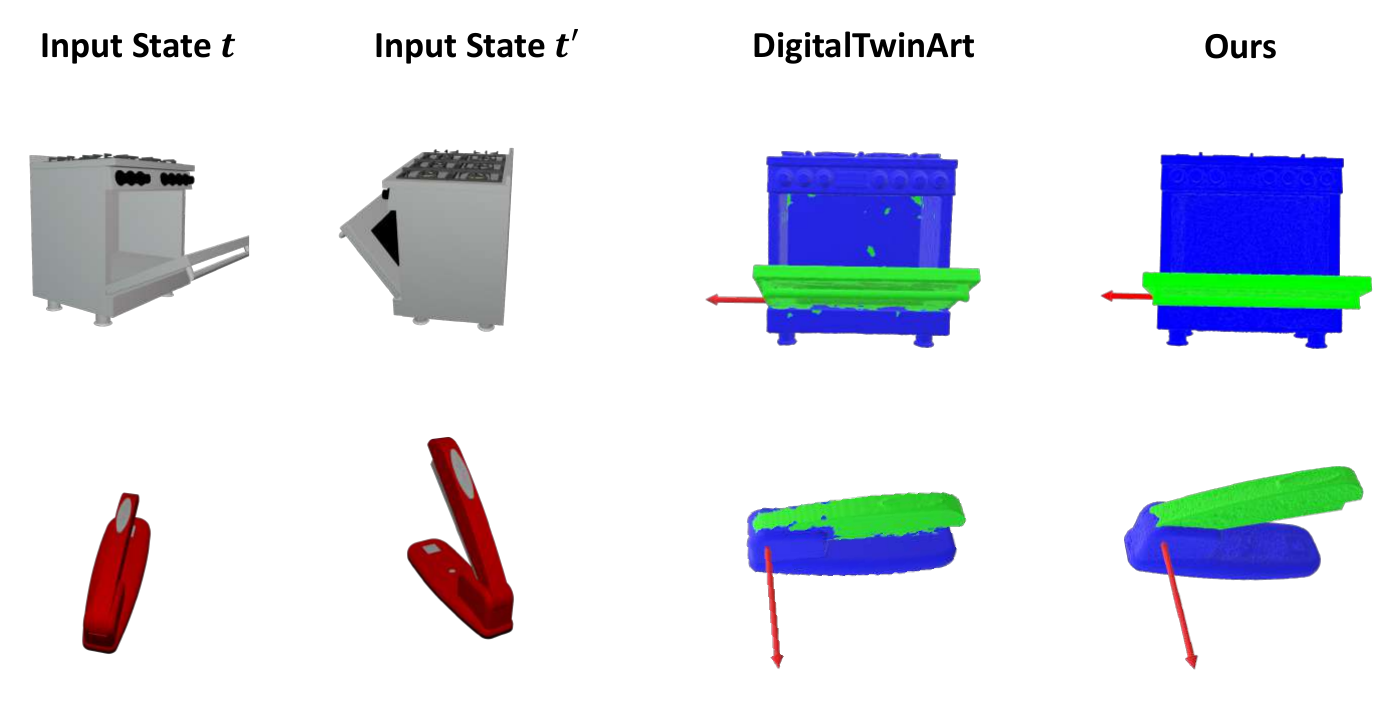}}
\caption{Qualitative analysis of shape reconstruction, part segmentation, and joint prediction results on the PARIS dataset. 
}
\label{fig:qualitative_paris2}
\end{figure*}

\begin{figure*}[th]
\centering
\noindent\makebox[1\textwidth][c]{\includegraphics[scale=0.55]{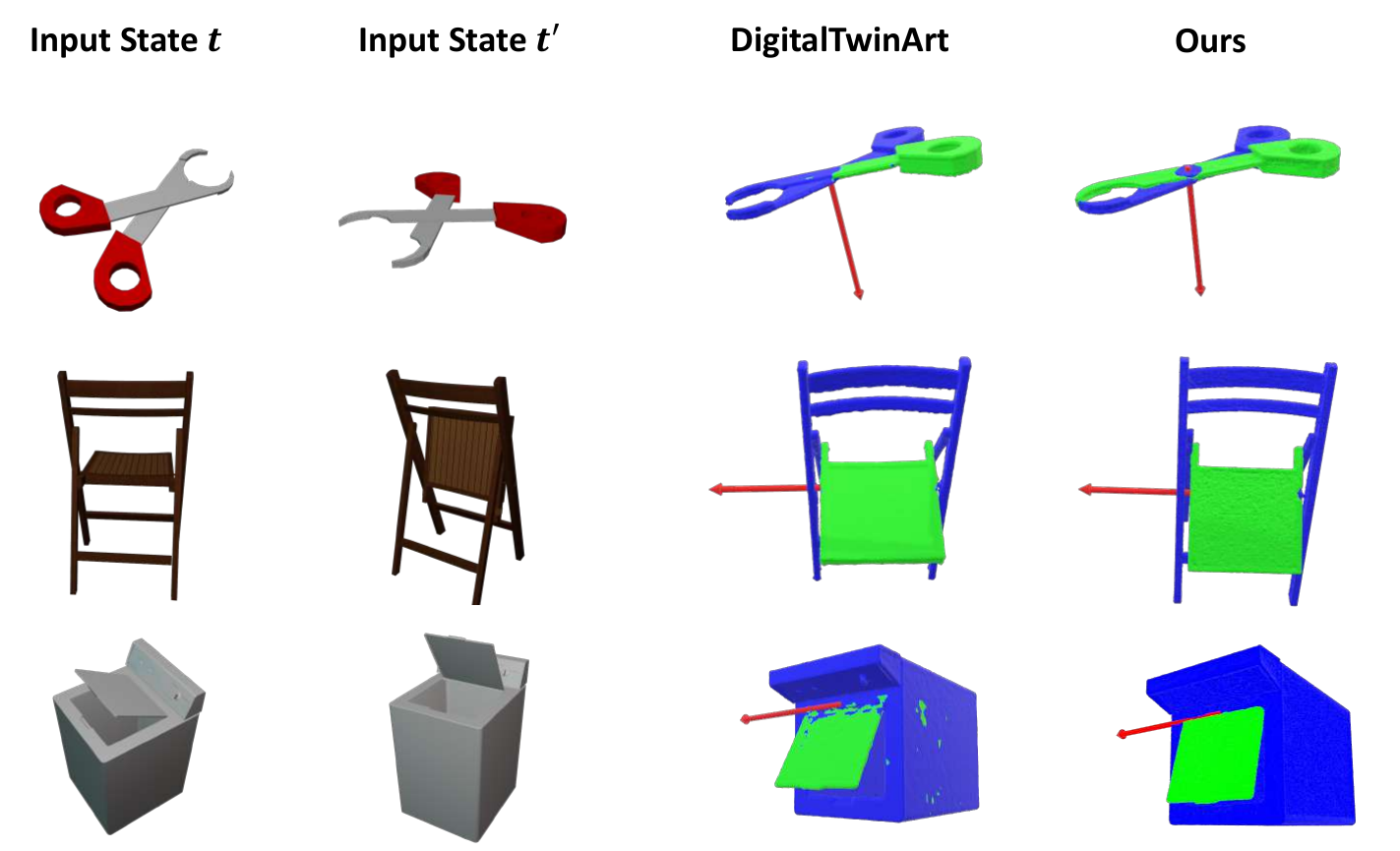}}
\caption{Qualitative analysis of shape reconstruction, part segmentation, and joint prediction results on the PARIS dataset. 
}
\label{fig:qualitative_paris3}
\end{figure*}


\begin{figure*}[th]
\centering
\noindent\makebox[1.0\textwidth][c]{\includegraphics[scale=0.6]{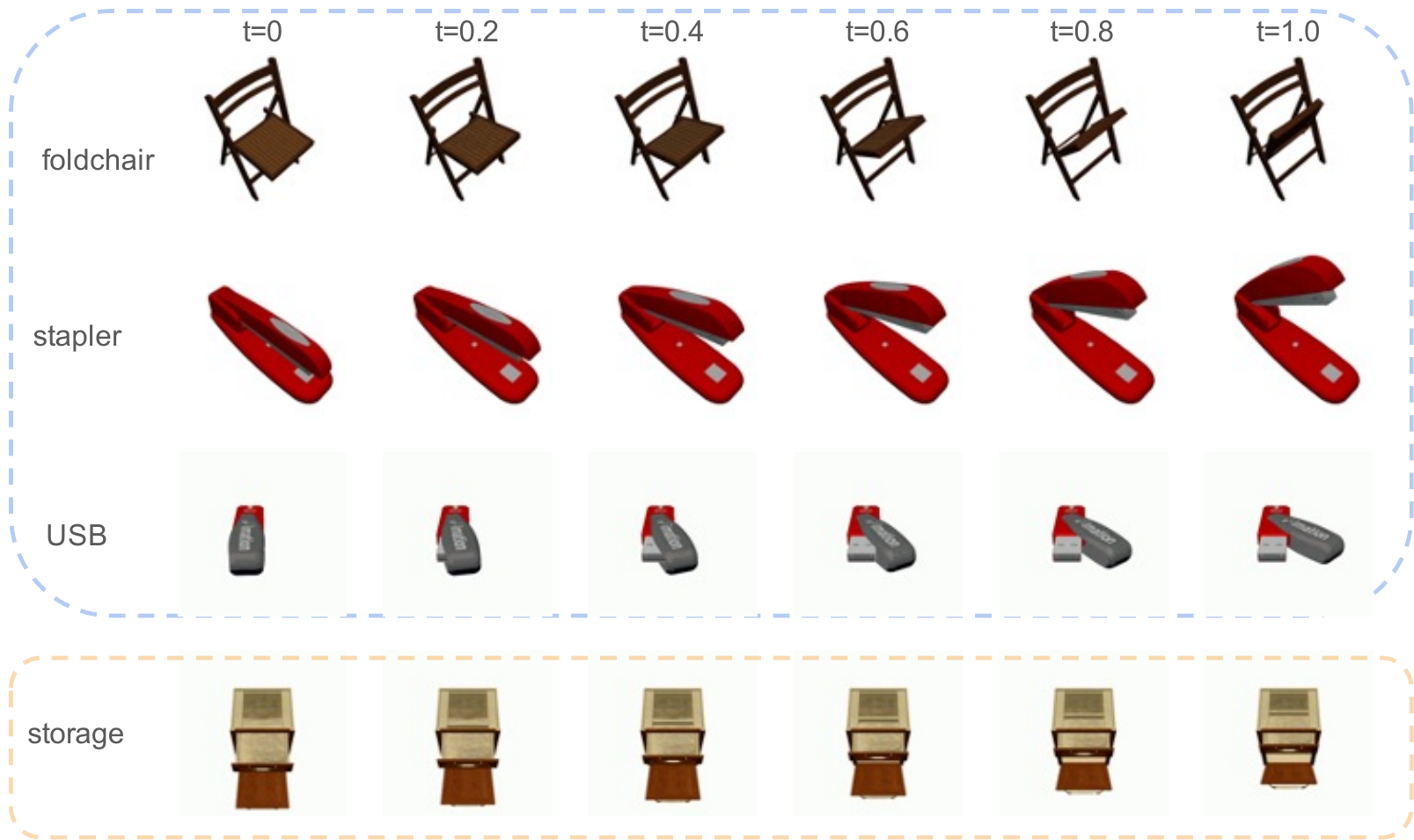}}
\caption{Qualitative analysis of scene interpolation on \textcolor{blue}{PARIS} and \textcolor{orange}{PartNet-Mobility} datasets.}
\label{fig:qualitative_paris_partnet}
\end{figure*}

\begin{figure*}[th]
\centering
\noindent\makebox[1.0\textwidth][c]{\includegraphics[scale=0.48]{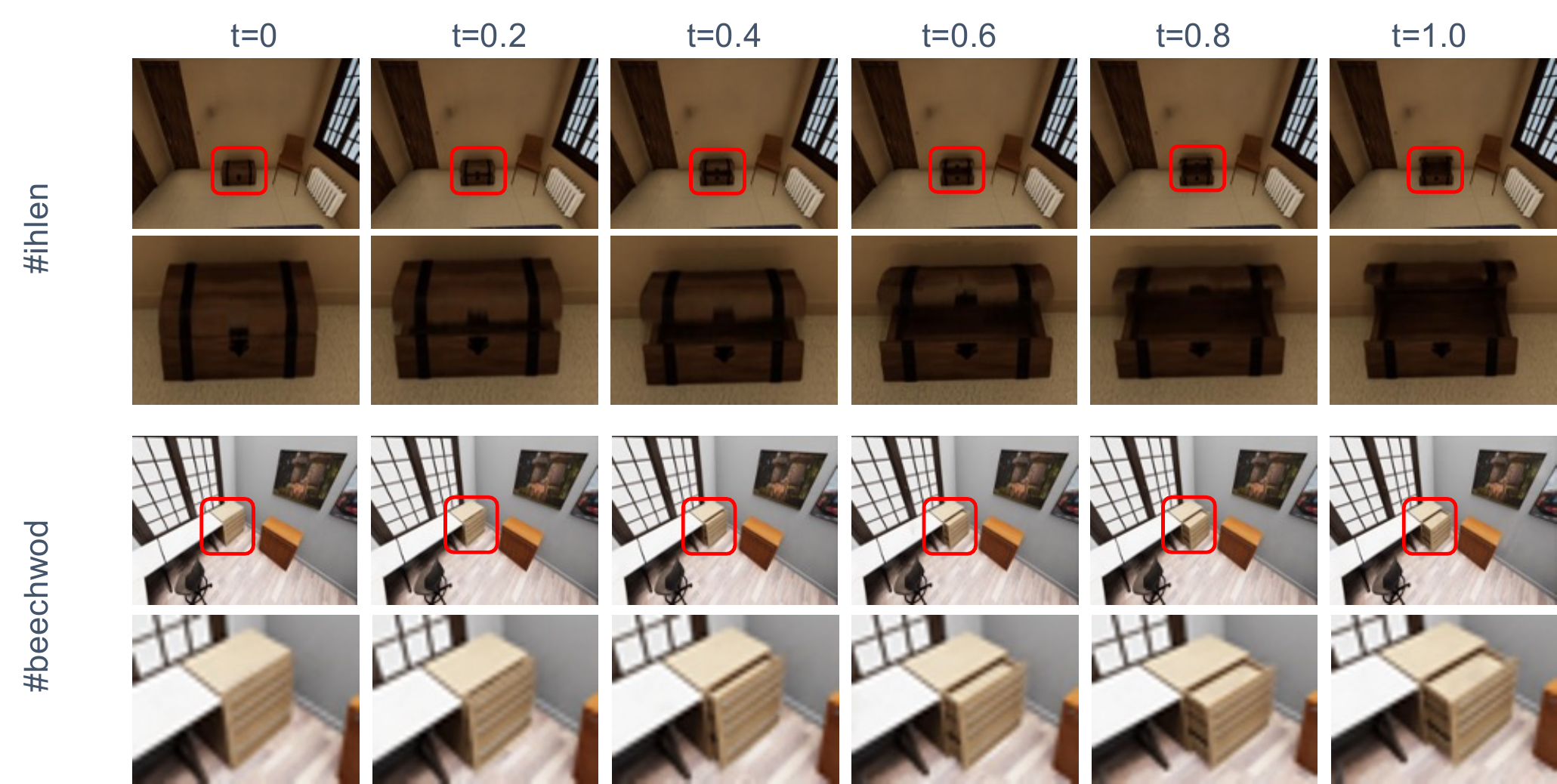}}
\caption{Motion snapshots on \textit{\#ihlen} and \textit{\#beechwod} from Indoor scene OmniSim dataset.}
\label{fig:qualitative_omnisim1}
\end{figure*}

\begin{figure*}[th]
\centering
\noindent\makebox[1.0\textwidth][c]{\includegraphics[scale=0.48]{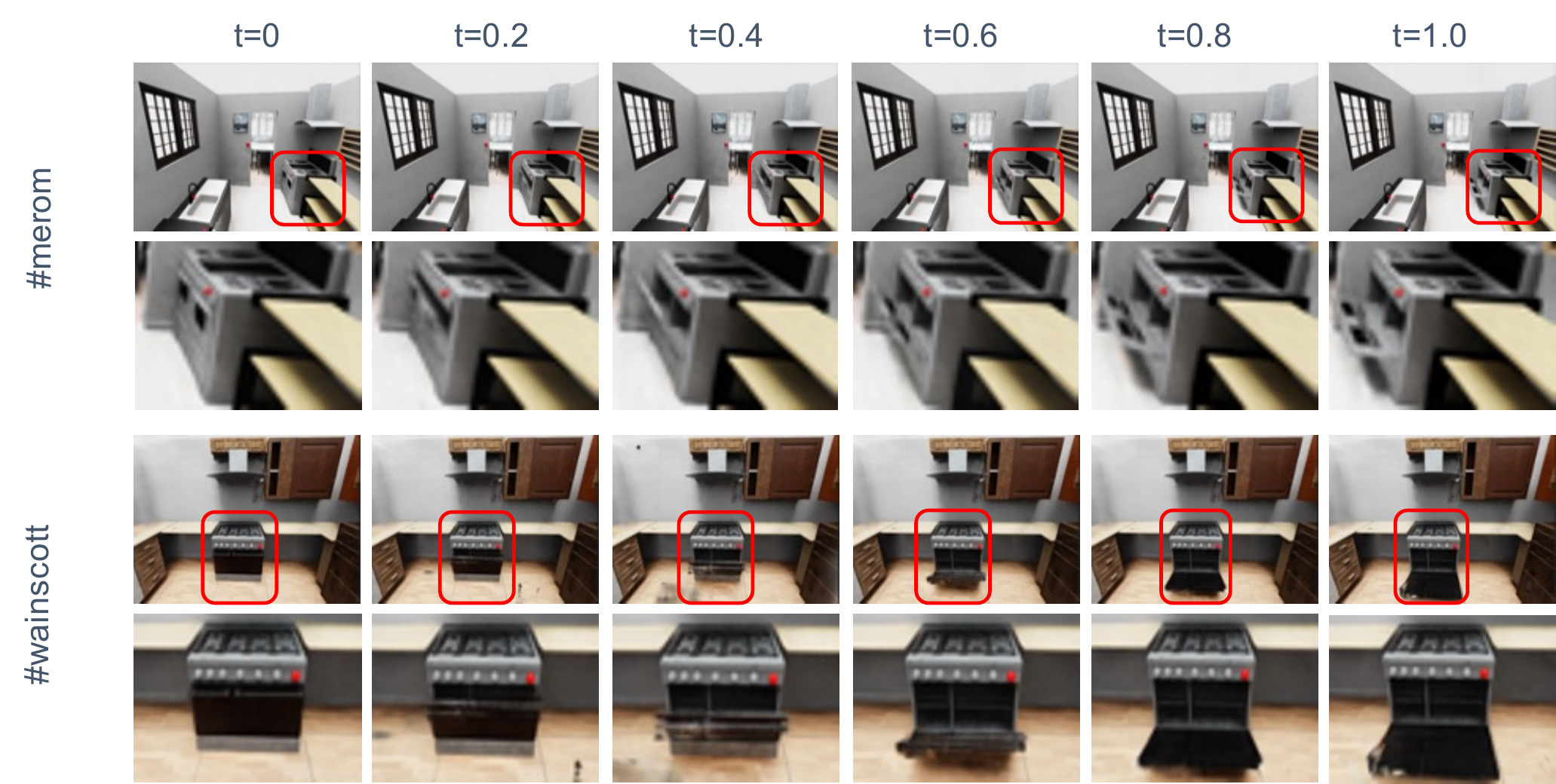}}
\caption{Motion snapshots on \textit{\#merom} and \textit{\#wainscott} from Indoor scene OmniSim dataset.}
\label{fig:qualitative_omnisim2}
\end{figure*}

\begin{figure*}[th]
\centering
\noindent\makebox[1\textwidth][c]{\includegraphics[scale=0.42]{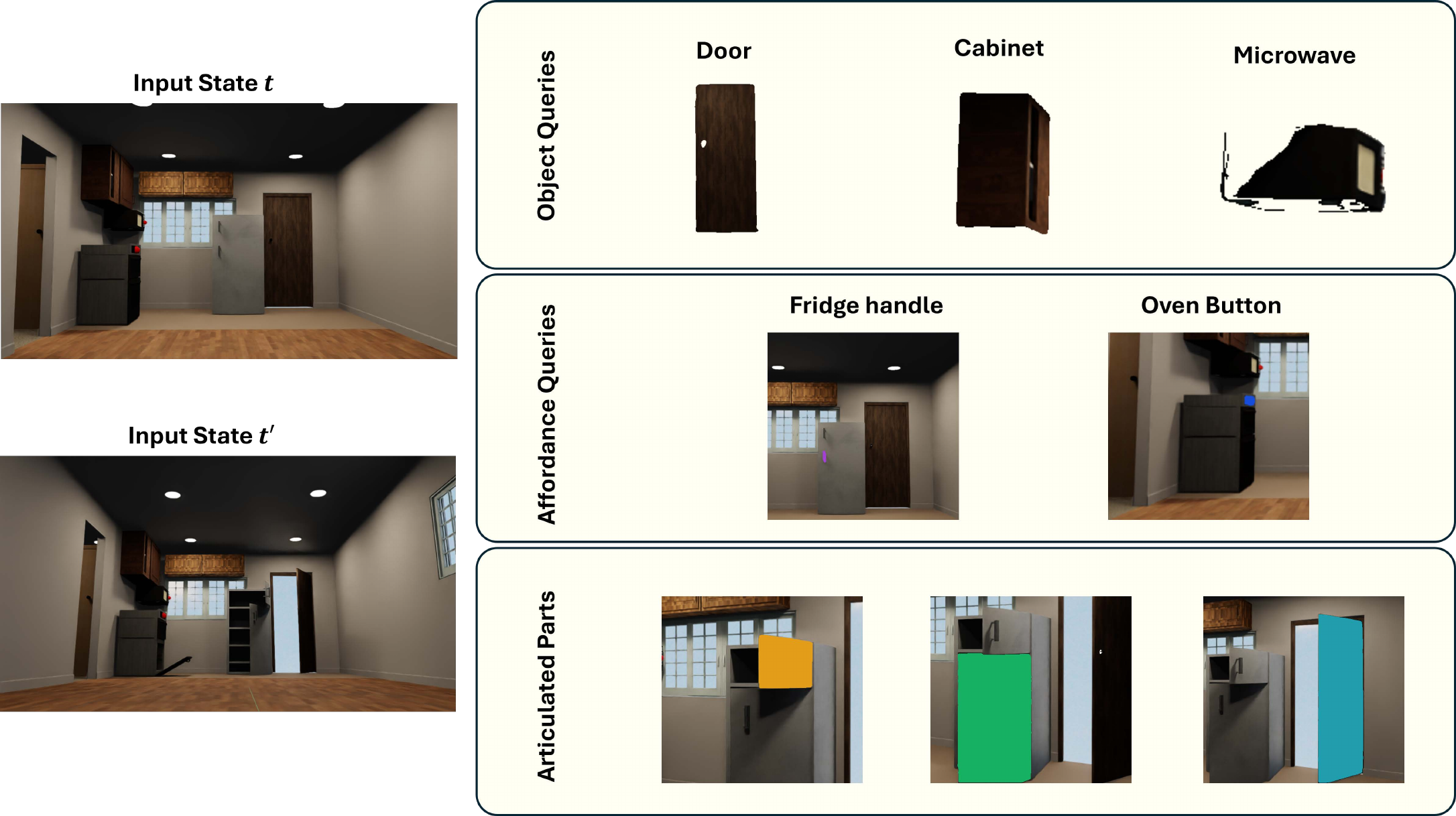}}
\caption{Qualitative analysis of object and affordance retrieval on one example scene from the Indoor scene OmniSim dataset.}
\label{fig:qualitative_omni}
\end{figure*}

\end{document}